\documentclass[aip,cha,reprint]{revtex4-1} 
\usepackage{lineno} 
\usepackage{graphicx} 
\usepackage{natbib} 
\usepackage{color} 
\usepackage{amssymb} 
\usepackage{subfigure} 
\usepackage{algorithm2e} 
\usepackage{dsfont} 

\usepackage[english]{babel}

\usepackage{amssymb,subfigure}

\begin{document} 


\title{Heterogeneous patterns enhancing static and dynamic texture classification} 
\author{N\'ubia Rosa da Silva} 
\affiliation{Scientific Computing Group \\ University of S\~{a}o Paulo \\ Instituto de Ci\^encias Matem\'aticas e de Computa\c{c}\~ao (ICMC) \\ http://scg.ifsc.usp.br \\ nubiars@icmc.usp.br} 
\author{Odemir Martinez Bruno} 
\affiliation{Scientific Computing Group \\ University of S\~{a}o Paulo \\ São Carlos Institute of Physics (IFSC) \\ http://scg.ifsc.usp.br \\ bruno@ifsc.usp.br}

\begin{abstract} 
Some mixtures, such as colloids like milk, blood, and gelatin, have homogeneous appearance when viewed with the naked eye, however, to observe them at the nanoscale is possible to understand the heterogeneity of its components. The same phenomenon can occur in pattern recognition in which it is possible to see heterogeneous patterns in texture images. However, current methods of texture analysis can not adequately describe such heterogeneous patterns. Common methods used by researchers analyse the image information in a global way, taking all its features in an integrated manner. Furthermore, multi-scale analysis verifies the patterns at different scales, but still preserving the homogeneous analysis. On the other hand various methods use textons to represent the texture, breaking texture down into its smallest unit. To tackle this problem, we propose a method to identify texture patterns not small as textons at distinct scales enhancing the separability among different types of texture. We find sub patterns of texture according to the scale and then group similar patterns for a more refined analysis. Tests were performed in four static texture databases and one dynamic one. Results show that our method provides better classification rate compared with conventional approaches both in static and in dynamic texture.
\end{abstract} 


\maketitle 

\section{Introduction}
\label{introduction}

Pattern recognition is the identification and interpretation of patterns in images, in order to extract relevant information on the image to identify and classify your content. Classification of patterns can be used in a variety of applications in different fields such as nanotechnology \cite{Florindo2012,Florindo2013,AnnampeduW2007,OhSP2011}, biology \cite{MesquitaSaJunior2013,Rossatto2011,Backes2011,GalasEW1985,IsasiZZ2011},  medicine \cite{LandeweerdGBH1981} and computer science \cite{Este2009,CanalsMR2010}. Different approaches have been developed according to the application, however, most of them analyze the information in a global way, using all the features in an integrated manner. 

One side in the pattern classification approach uses multi-scale analysis of patterns, that is, different scales of observation are used to perform the analysis and find similar patterns,  because important structures in an image usually occur at different spatial scales \cite{Koenderink1984} \cite{Witkin1983}. Methods based on textons \cite{Julesz1981,Leung1999} represent each pixel of a texture as the convolution of a multi-scale and multi-orientation filter bank  producing a texton vocabulary. Thus, these methods of texture analysis characterize the image homogeneously on the scale to be analyzed. Both the overall analysis, such as multi-scale approach is appropriate for the vast majority of problems in pattern recognition. However, in some problems, due to the heterogeneous nature of the composition of the objects under consideration, it is needed one more step in the process of pattern recognition, the analysis of heterogeneous patterns.

The aim of this paper is to demonstrate how can apply heterogeneous analysis to improve results regarding the homogeneous analysis. To validate our proposal experiments were performed on four static e one dynamic texture databases using the same texture descriptor but with different approaches. In all tests, heterogeneous analysis proved to be better than the homogeneous analysis enhancing the rate classification.

This paper is organized as follows: Section \ref{method} describes the heterogeneous pattern analysis. Section \ref{results} shows the results and discussions and in Section \ref{conclusions} the conclusions.

\section{Method}
\label{method}

\begin{figure*}[!htbp]
    \center
    \subfigure[]{
    	\includegraphics[height=4cm]{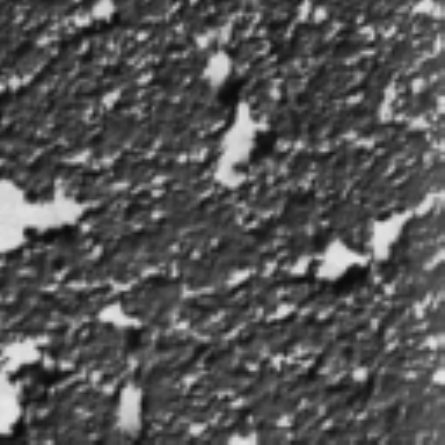} 	
    	\label{fig:2patterns}}
    \hfil
    \subfigure[]{
    	\includegraphics[height=4cm]{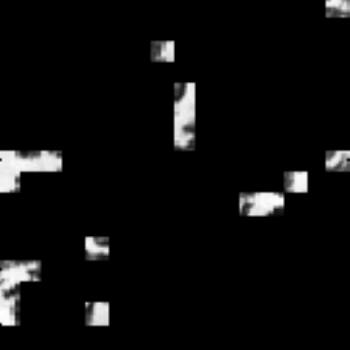}
    	\label{fig:segmentaTexturaa}}
    \hfil
	\subfigure[]{
		\includegraphics[height=4cm]{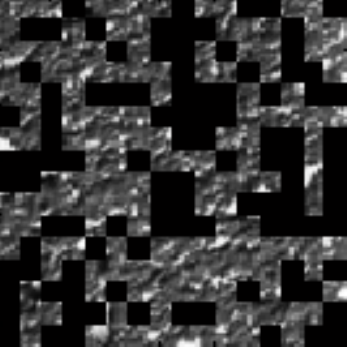}
	    \label{fig:segmentaTexturab}}
    \caption{Segmentation according to the heterogeneous patterns. (a) Original image. (b) and (c) Two patterns obtained from (a).}
\end{figure*}

Heterogeneous pattern arises when the object under analysis presents combined patterns in its composition.
Similar to the classic definition of heterogeneous compositions in chemical compounds \cite{Staley2004}, which is the characteristic of presenting a different appearance or composition when analyzed in parts. 
The same analogy can be applied in recognizing patterns in images. In each image can be found heterogeneous patterns, i. e., it is possible to distinguish the different patterns in each of the texture images. Figure \ref{fig:2patterns} shows an image from Brodatz, which analyzed by conventional methods, would be defined only a pattern for this image. Using the approach of heterogeneous patterns, two different types of texture patterns are identified in its formation.

This new view to analyze the various patterns requires a new approach for analysis of similarity between images. The first step is to segment the texture by defining regions where the image belong to a given pattern while defining the patterns in the image. Figures \ref{fig:segmentaTexturaa} and \ref{fig:segmentaTexturab} illustrate the segmentation of the image of Figure \ref{fig:2patterns} in two patterns.

To obtain this result the image was divided into smaller sized windows of  8 $\times$ 8 pixels, for windows with texture sufficiently homogeneous, that is, where only one pattern is found. Furthermore, there was obtained a characteristic to represent each window. Windows with similar characteristics remain the same pattern, and those with distinct characteristics were separated into different patterns. Figures \ref{fig:segmentaTexturaa} and \ref{fig:segmentaTexturab} show the windows stayed grouped using \textit{k-means} algorithm according to their pattern with the remaining windows in each pattern. Textural features from Haralick \cite{Haralick1973} descriptors were used to characterize each window. 

Haralick  descriptors \cite{Haralick1973} are based on the spatial gray level dependence matrices, or co-occurrence matrix. Contrast, Correlation, Energy and Homogeneity were computed from resulting co-occurrence matrices to obtain a set of 32 descriptors for each window. Let $G$ be the gray levels and $p(i,j)$ a matrix of relative frequencies of two neighboring resolution cell with intensity $i$ and $j$ separated by distance $d$ and direction $\theta$. Extracted features are described as follows:

\begin{equation}
Contrast = \sum_{i=0}^{G-1} \sum_{j=0}^{G-1} (i-j)^2 p(i,j)
\end{equation}

\begin{equation}
Correlation = \sum_{i=0}^{G-1} \sum_{j=0}^{G-1}  \frac{ijp(i,j) - \mu _x \mu _y}{\sigma_x\sigma_y}   
\end{equation}

\begin{equation}
Energy =  \sum_{i=0}^{G-1} \sum_{j=0}^{G-1} p(i,j)^2 
\end{equation}

\begin{equation}
Homogeneity =  \sum_{i=0}^{G-1} \sum_{j=0}^{G-1}  \frac{p(i,j)}{1+|i-j|}
\end{equation}

$\mu _x, \mu _y, \sigma_x$ and $\sigma_y$ are,  respectively, means and standard deviations of the sum of elements of each row and column of the co-occurrence matrix. Distances of 1 and 2 pixels with angles of $0\,^{\circ}$, $45\,^{\circ}$, $90\,^{\circ}$ and $135\,^{\circ}$ were used.

\begin{figure*}[!htbp]
    \centering
    \subfigure[]{\includegraphics[height=3.5cm]{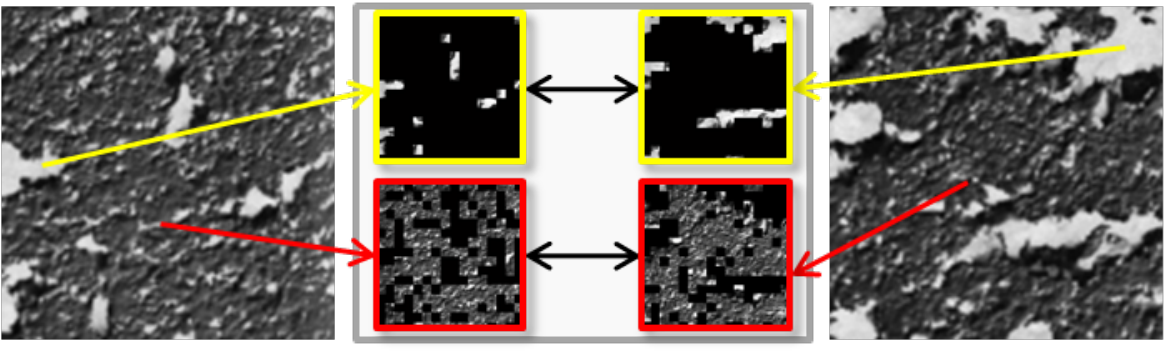}
    		     \label{fig:matchinga} }
	\subfigure[]{\includegraphics[height=3.5cm]{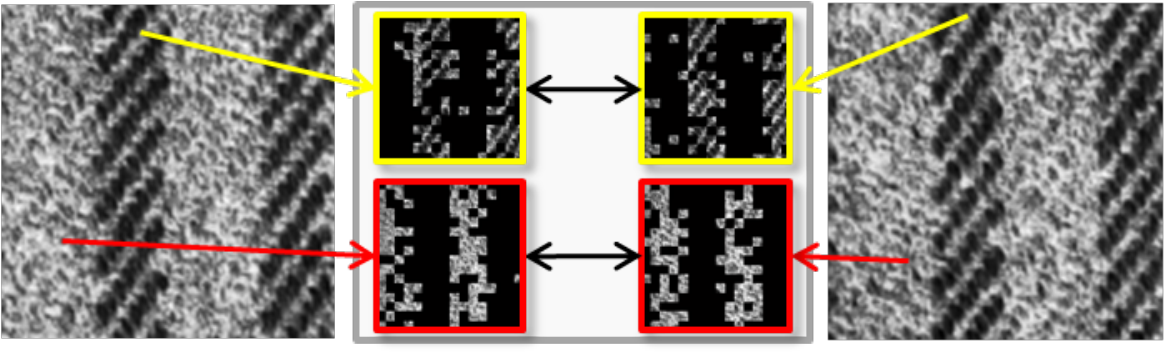}
		    \label{fig:matchingb}}
	\subfigure[]{\includegraphics[height=3.5cm]{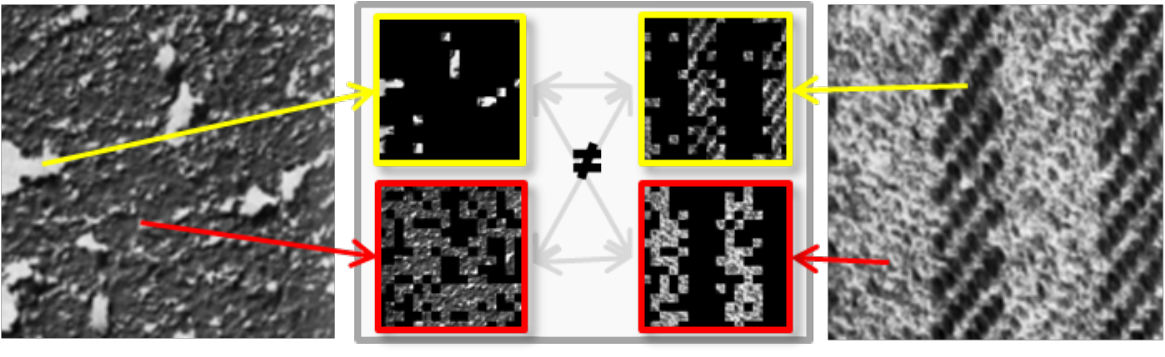}
			\label{fig:nomatching}}
    \caption{Matching patterns. (a) and (b) matching of two images from the same class. (c) Matching of patterns can not be made because the images belong to different classes.}
\end{figure*}

After the windows were divided into patterns, we performed a survival analysis of windows, 25\% of the windows with features farthest from the group average were discarded making the representation of pattern more consistent. Each pattern was characterized by a feature vector average that is the average of the characteristics of all windows belonging to the pattern. The next step is to find the best matching among the patterns to define the image similarity as Figures \ref{fig:matchinga} and \ref{fig:matchingb} exemplify. They show two examples where it has an optimal matching among the image patterns of the same class. When  the best fit among all the patterns in each image is found, we have the degree of similarity, defining whether the images belong to the same class. Figure \ref{fig:nomatching} shows an example where it is not possible to perform the fitting patterns and all possibilities of fitting are tested. In this case the images belong to different classes.

\begin{figure*}[!htbp]
    \centering
    \includegraphics[height=4cm]{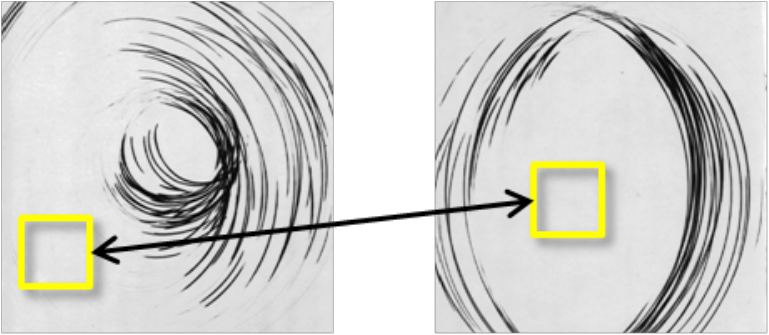}
    \caption{Same pattern in different classes.}
    \label{fig:samepattern}
\end{figure*}

However, this approach has a special feature when the same pattern can be found in different classes of texture (See Figure \ref{fig:samepattern}). This one influences the analysis of the similarity degree between classes, because different classes can obtain a high rate of similarity, since some patterns will have great matching.

\section{Results and Discussions}
\label{results}

\begin{figure*}[!htbp]
    \centering
    \subfigure[]{
    	\includegraphics[height=4cm, width=4cm]{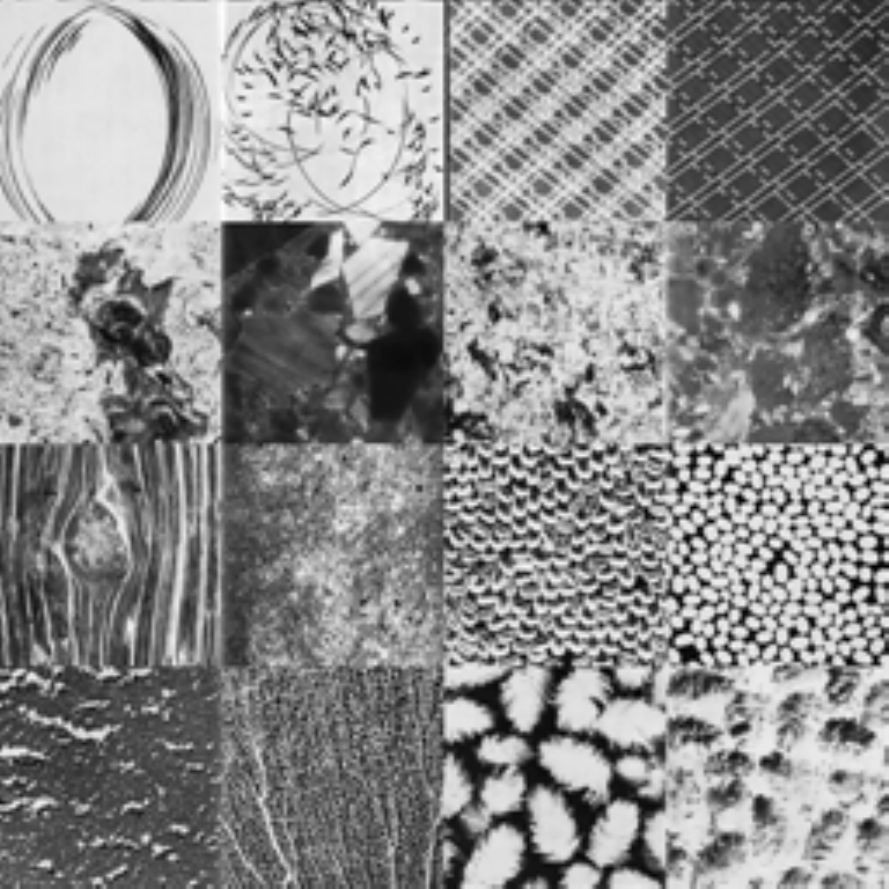}	
    	\label{fig:brodatz}}
    \subfigure[]{
    	\includegraphics[height=4cm, width=4cm]{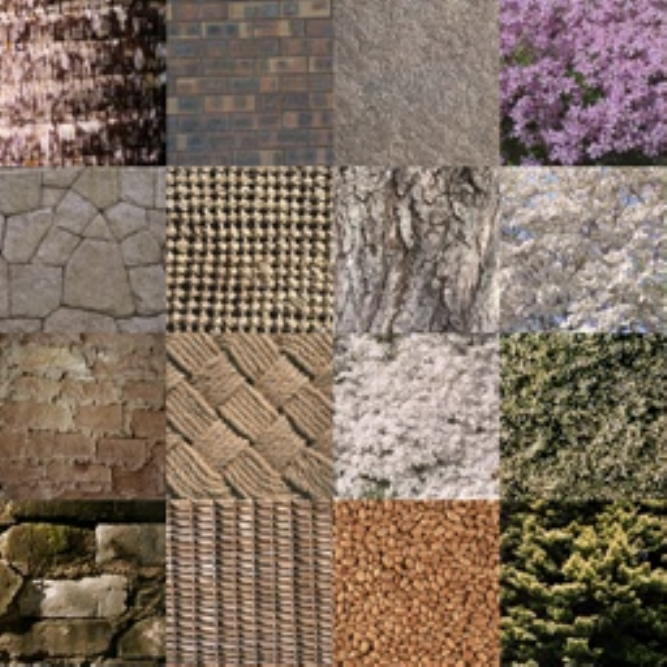}
    	\label{fig:vistex}}
	\subfigure[]{
		\includegraphics[height=4cm, width=4cm]{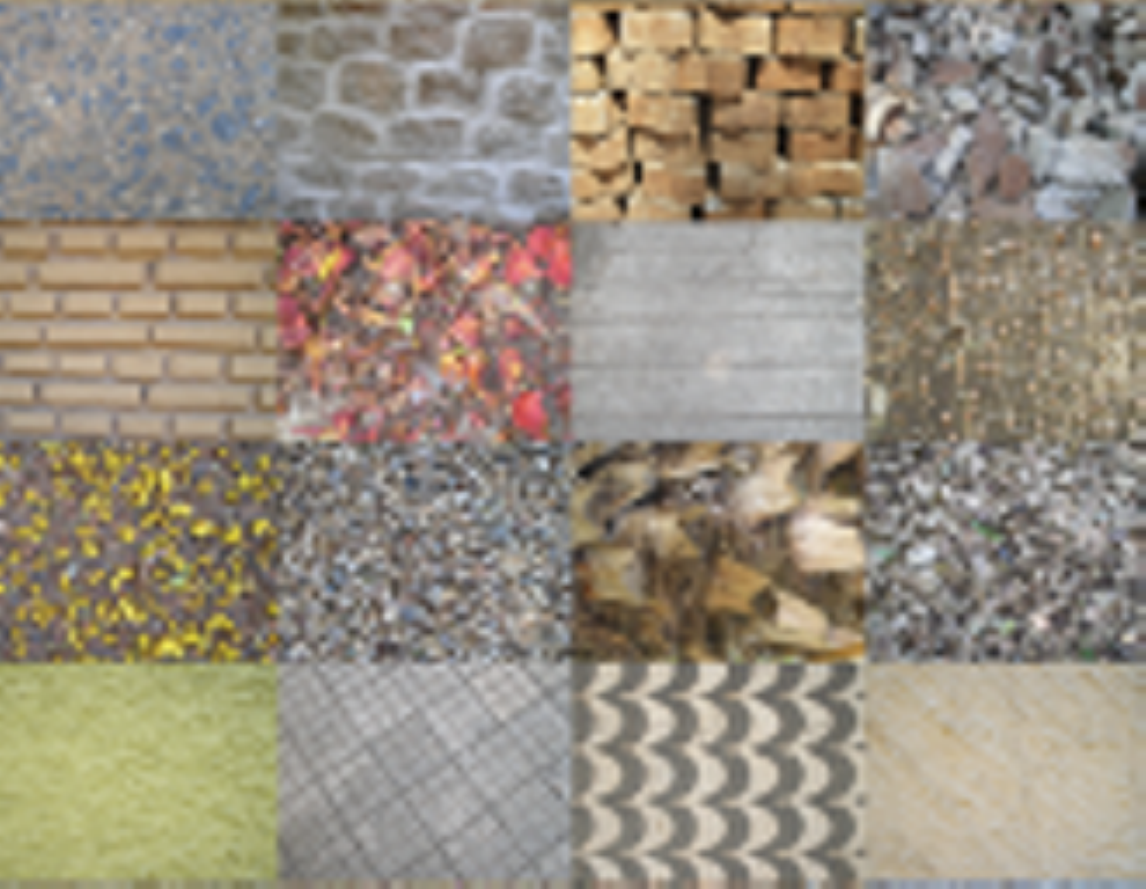}
	    \label{fig:usptex}}
	\subfigure[]{
		\includegraphics[height=4cm, width=4cm]{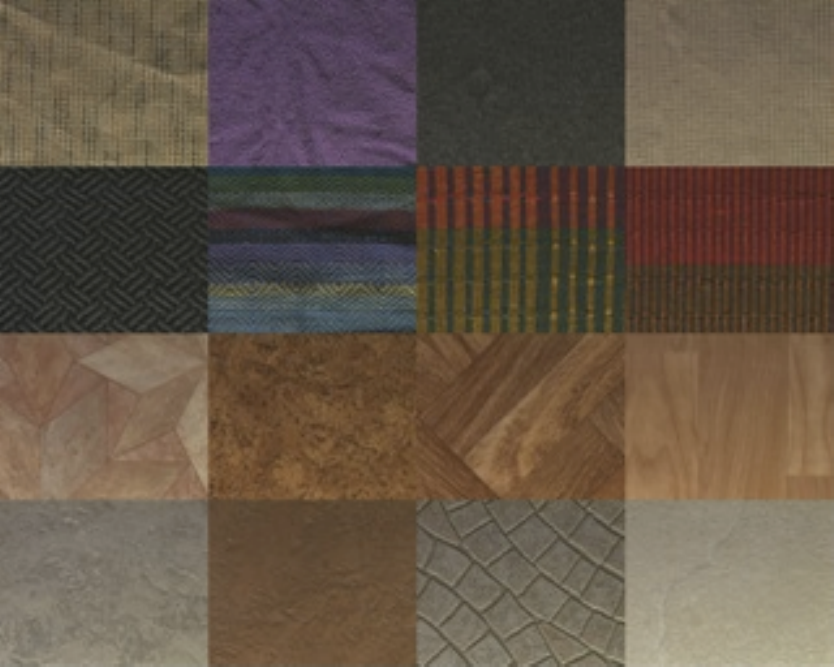}
	    \label{fig:outex}}
    \caption{Samples of (a) Brodatz, (b) Vistex, (c) Usptex and (d) Outex.}
    \label{fig:databases}
\end{figure*}

\begin{figure*}[!htbp]
    \centering
    \includegraphics[height=6cm]{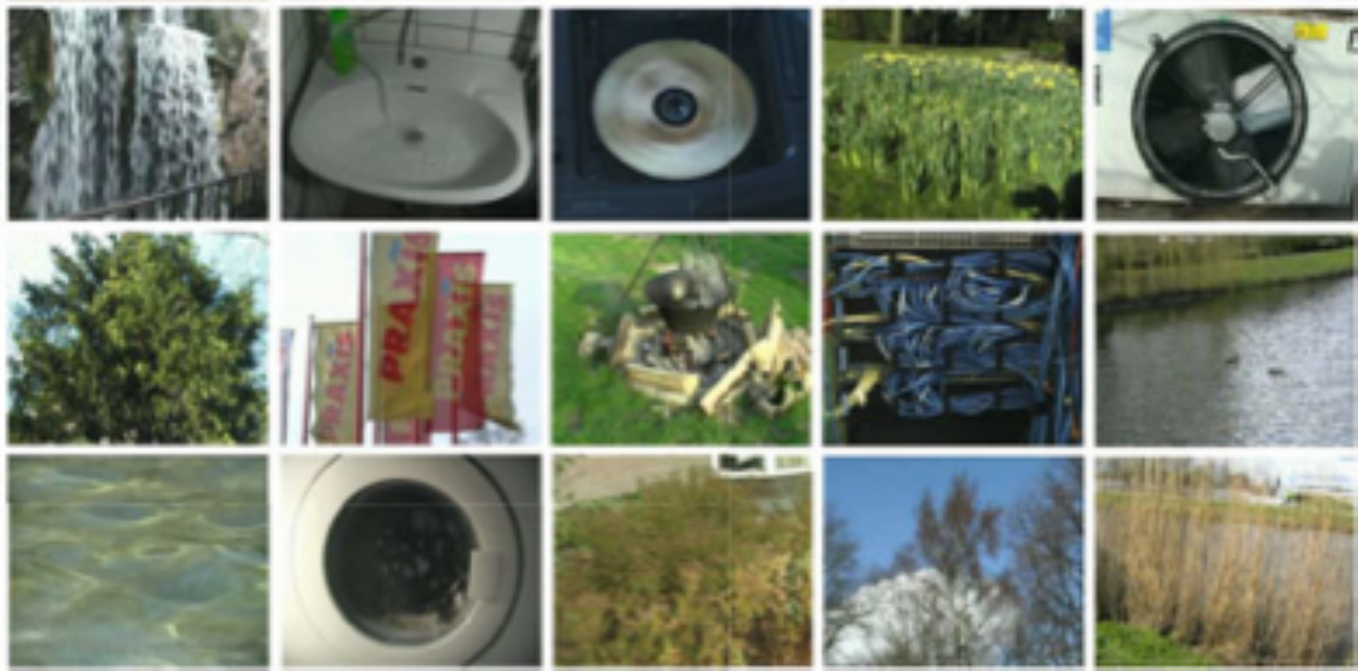}
    \caption{Samples images from Dyntex, a dynamic texture database.}
    \label{fig:dyntex}
\end{figure*}

To validate the method  of heterogeneous patterns analysis, this technique was applied in four different texture databases, each with its peculiarities: Brodatz, USPTex, Vistex and  Outex and one dynamic texture database: Dyntex. Figure \ref{fig:databases} shows some samples of the texture databases and Figure \ref{fig:dyntex} shows examples of dynamic texture database Dyntex. The classification method  $k$-Nearest Neighbor ($k-NN$) with 10-fold cross-validation scheme was used in all experiments. 

\begin{description}

\item[Brodatz] \cite{Brodatz1966} contains  1110 natural textures of 200 $\times$ 200 pixels divided into 111 classes (Figure \ref{fig:brodatz}).

\item[Vistex] \cite{vistex2009} contains  864 images of 128 $\times$ 128 pixels size with 54 texture classes (Figure \ref{fig:vistex}). 

\item[USPTex] \cite{usptex2012} has 3984 natural texture images of 128 $\times$ 128 pixels size divided into 332 classes (Figure \ref{fig:usptex}). 

\item[Outex] \cite{Ojala2002} has 1360 images of 128 $\times$ 128 pixels size divided into 68 classes (Figure \ref{fig:outex}).

\item[Dyntex] \cite{DyntexPeteri2010} consists of 1230 videos with 250 frames with 400 $\times$ 300 pixels size divided into 123 classes of dynamic texture (Figure \ref{fig:dyntex}).

\end{description}

The method used as classical approach was Haralick descriptors with the same configurations of characterization of windows. This way we can compare the same descriptor but with different perspectives. One of the most important parameters is the window size 
that directly interferes in the results. Large windows remain with the general representation of the image, preserving the homogeneous analysis of patterns. However, becomes more homogeneous window, or small windows, the classification rate increases, as the representative pattern of the window increases. For this experiment two patterns were analyzed in each image. Table \ref{tab:res2clusters} shows the results obtained when using the conventional method in comparison with heterogeneous patterns, in all cases heterogeneous pattern method improved the classification rate proving to be better than the standard approach in which only one pattern is established for the entire image. 

\begin{table*}[!htbp]
\begin{center}
\caption{Accuracy rate by comparing the traditional method of analysis with the heterogeneous method.}
\label{tab:res2clusters}
\begin{tabular}{lrrrrr}
\hline
Database 			          & Brodatz & \,\,Vistex & Usptex & \,\,Outex & Dyntex  \\
\hline
Classical approach          & 88.92   &	89.47  & 69.58	& 76.62 & 76.89 \\
Heterogeneous pattern & 93.06   &	93.63  & 78.49	& 76.99 & 98.29 \\
\hline
\end{tabular}
\end{center}
\end{table*}

\section{Conclusions}
\label{conclusions}

Usually, analysis in texture pattern recognition using the homogeneous approach taking all the image information in a global way, that is, the entire image is defined as one pattern. However, it is possible to find more than one texture pattern on the image that we called here, heterogeneous pattern. It is necessary to map the patterns in each image and then check the similarity of patterns among them. This method improved the accuracy rate of classification using the same method to characterize the image showing robustness evaluating heterogeneous patterns on images.

\section*{Acknowledgments}

N\'ubia Rosa da Silva acknowledges support from FAPESP (2011/21467-9) and Odemir Martinez Bruno acknowledges the financial support of CNPq (308449/2010-0 and 473893/2010-0) and FAPESP (2011/01523-1).


\begin{thebibliography}{23}%
\makeatletter
\providecommand \@ifxundefined [1]{%
 \@ifx{#1\undefined}
}%
\providecommand \@ifnum [1]{%
 \ifnum #1\expandafter \@firstoftwo
 \else \expandafter \@secondoftwo
 \fi
}%
\providecommand \@ifx [1]{%
 \ifx #1\expandafter \@firstoftwo
 \else \expandafter \@secondoftwo
 \fi
}%
\providecommand \natexlab [1]{#1}%
\providecommand \enquote  [1]{``#1''}%
\providecommand \bibnamefont  [1]{#1}%
\providecommand \bibfnamefont [1]{#1}%
\providecommand \citenamefont [1]{#1}%
\providecommand \href@noop [0]{\@secondoftwo}%
\providecommand \href [0]{\begingroup \@sanitize@url \@href}%
\providecommand \@href[1]{\@@startlink{#1}\@@href}%
\providecommand \@@href[1]{\endgroup#1\@@endlink}%
\providecommand \@sanitize@url [0]{\catcode `\\12\catcode `\$12\catcode
  `\&12\catcode `\#12\catcode `\^12\catcode `\_12\catcode `\%12\relax}%
\providecommand \@@startlink[1]{}%
\providecommand \@@endlink[0]{}%
\providecommand \url  [0]{\begingroup\@sanitize@url \@url }%
\providecommand \@url [1]{\endgroup\@href {#1}{\urlprefix }}%
\providecommand \urlprefix  [0]{URL }%
\providecommand \Eprint [0]{\href }%
\@ifxundefined \urlstyle {%
  \providecommand \doi  [0]{\begingroup \@sanitize@url \@doi}%
  \providecommand \@doi [1]{\endgroup \@@startlink {\doibase
  #1}doi:\discretionary {}{}{}#1\@@endlink }%
}{%
  \providecommand \doi  [0]{doi:\discretionary{}{}{}\begingroup
  \urlstyle{rm}\Url }%
}%
\providecommand \doibase [0]{http://dx.doi.org/}%
\providecommand \Doi [0]{\begingroup \@sanitize@url \@Doi }%
\providecommand \@Doi  [1]{\endgroup\@@startlink{\doibase#1}\@@Doi}%
\providecommand \@@Doi [1]{#1\@@endlink}%
\providecommand \selectlanguage [0]{\@gobble}%
\providecommand \bibinfo  [0]{\@secondoftwo}%
\providecommand \bibfield  [0]{\@secondoftwo}%
\providecommand \translation [1]{[#1]}%
\providecommand \BibitemOpen [0]{}%
\providecommand \bibitemStop [0]{}%
\providecommand \bibitemNoStop [0]{.\EOS\space}%
\providecommand \EOS [0]{\spacefactor3000\relax}%
\providecommand \BibitemShut  [1]{\csname bibitem#1\endcsname}%
\bibitem [{\citenamefont {Florindo}\ \emph {et~al.}(2012)\citenamefont
  {Florindo}, \citenamefont {Sikora}, \citenamefont {Pereira},\ and\
  \citenamefont {Bruno}}]{Florindo2012}%
  \BibitemOpen
  \bibfield  {author} {\bibinfo {author} {\bibfnamefont {J.~B.}\ \bibnamefont
  {Florindo}}, \bibinfo {author} {\bibfnamefont {M.~S.}\ \bibnamefont
  {Sikora}}, \bibinfo {author} {\bibfnamefont {E.~C.}\ \bibnamefont {Pereira}},
  \ and\ \bibinfo {author} {\bibfnamefont {O.~M.}\ \bibnamefont {Bruno}},\
  }\bibfield  {title} {\enquote {\bibinfo {title} {Multiscale fractal
  descriptors applied to nanoscale images},}\ }\href@noop {} {\bibfield
  {journal} {\bibinfo  {journal} {Journal of Superconductivity and Novel
  Magnetism},\ \bibinfo {pages} {1--6}} (\bibinfo {year} {2012})}\BibitemShut
  {NoStop}%
\bibitem [{\citenamefont {Florindo}\ \emph {et~al.}(2013)\citenamefont
  {Florindo}, \citenamefont {Sikora}, \citenamefont {Pereira},\ and\
  \citenamefont {Bruno}}]{Florindo2013}%
  \BibitemOpen
  \bibfield  {author} {\bibinfo {author} {\bibfnamefont {J.~B.}\ \bibnamefont
  {Florindo}}, \bibinfo {author} {\bibfnamefont {M.~S.}\ \bibnamefont
  {Sikora}}, \bibinfo {author} {\bibfnamefont {E.~C.}\ \bibnamefont {Pereira}},
  \ and\ \bibinfo {author} {\bibfnamefont {O.~M.}\ \bibnamefont {Bruno}},\
  }\bibfield  {title} {\enquote {\bibinfo {title} {Characterization of
  nanostructured material images using fractal descriptors},}\ }\href@noop {}
  {\bibfield  {journal} {\bibinfo  {journal} {Physica A: Statistical Mechanics
  and its Applications},\ }\textbf {\bibinfo {volume} {392}},\ \bibinfo {pages}
  {1694--1701} (\bibinfo {year} {2013})},\ ISSN \bibinfo {issn}
  {0378-4371}\BibitemShut {NoStop}%
\bibitem [{\citenamefont {Annampedu}\ and\ \citenamefont
  {Wagh}(2007)}]{AnnampeduW2007}%
  \BibitemOpen
  \bibfield  {author} {\bibinfo {author} {\bibfnamefont {V.}~\bibnamefont
  {Annampedu}}\ and\ \bibinfo {author} {\bibfnamefont {M.~D.}\ \bibnamefont
  {Wagh}},\ }\bibfield  {title} {\enquote {\bibinfo {title} {Reconfigurable
  approximate pattern matching architectures for nanotechnology},}\ }\Doi
  {10.1016/j.mejo.2007.01.020} {\bibfield  {journal} {\bibinfo  {journal}
  {Microelectronics Journal},\ }\textbf {\bibinfo {volume} {38}},\ \bibinfo
  {pages} {430 -- 438} (\bibinfo {year} {2007})},\ ISSN \bibinfo {issn}
  {0026-2692}\BibitemShut {NoStop}%
\bibitem [{\citenamefont {Oh}\ \emph {et~al.}(2011)\citenamefont {Oh},
  \citenamefont {Song},\ and\ \citenamefont {Park}}]{OhSP2011}%
  \BibitemOpen
  \bibfield  {author} {\bibinfo {author} {\bibfnamefont {E.~H.}\ \bibnamefont
  {Oh}}, \bibinfo {author} {\bibfnamefont {H.~S.}\ \bibnamefont {Song}}, \ and\
  \bibinfo {author} {\bibfnamefont {T.~H.}\ \bibnamefont {Park}},\ }\bibfield
  {title} {\enquote {\bibinfo {title} {Recent advances in electronic and
  bioelectronic noses and their biomedical applications},}\ }\Doi
  {10.1016/j.enzmictec.2011.04.003} {\bibfield  {journal} {\bibinfo  {journal}
  {Enzyme and Microbial Technology},\ }\textbf {\bibinfo {volume} {48}},\
  \bibinfo {pages} {427 -- 437} (\bibinfo {year} {2011})},\ ISSN \bibinfo
  {issn} {0141-0229}\BibitemShut {NoStop}%
\bibitem [{\citenamefont {de~Mesquita S\'a~Junior}\ \emph
  {et~al.}(2013)\citenamefont {de~Mesquita S\'a~Junior}, \citenamefont
  {Rossatto}, \citenamefont {Kolb},\ and\ \citenamefont
  {Bruno}}]{MesquitaSaJunior2013}%
  \BibitemOpen
  \bibfield  {author} {\bibinfo {author} {\bibfnamefont {J.~J.}\ \bibnamefont
  {de~Mesquita S\'a~Junior}}, \bibinfo {author} {\bibfnamefont {D.~R.}\
  \bibnamefont {Rossatto}}, \bibinfo {author} {\bibfnamefont {R.~M.}\
  \bibnamefont {Kolb}}, \ and\ \bibinfo {author} {\bibfnamefont {O.~M.}\
  \bibnamefont {Bruno}},\ }\bibfield  {title} {\enquote {\bibinfo {title} {A
  computer vision approach to quantify leaf anatomical plasticity: a case study
  on gochnatia polymorpha (less.) cabrera},}\ }\Doi
  {10.1016/j.ecoinf.2013.02.007} {\bibfield  {journal} {\bibinfo  {journal}
  {Ecological Informatics},\ }\textbf {\bibinfo {volume} {15}},\ \bibinfo
  {pages} {34 -- 43} (\bibinfo {year} {2013})},\ ISSN \bibinfo {issn}
  {1574-9541}\BibitemShut {NoStop}%
\bibitem [{\citenamefont {Rossatto}\ \emph {et~al.}(2011)\citenamefont
  {Rossatto}, \citenamefont {Casanova}, \citenamefont {Kolb},\ and\
  \citenamefont {Bruno}}]{Rossatto2011}%
  \BibitemOpen
  \bibfield  {author} {\bibinfo {author} {\bibfnamefont {D.}~\bibnamefont
  {Rossatto}}, \bibinfo {author} {\bibfnamefont {D.}~\bibnamefont {Casanova}},
  \bibinfo {author} {\bibfnamefont {R.}~\bibnamefont {Kolb}}, \ and\ \bibinfo
  {author} {\bibfnamefont {O.~M.}\ \bibnamefont {Bruno}},\ }\bibfield  {title}
  {\enquote {\bibinfo {title} {Fractal analysis of leaf-texture properties as a
  tool for taxonomic and identification purposes: a case study with species
  from neotropical melastomataceae (miconieae tribe)},}\ }\href@noop {}
  {\bibfield  {journal} {\bibinfo  {journal} {Plant Systematics and
  Evolution},\ }\textbf {\bibinfo {volume} {291}},\ \bibinfo {pages} {103--116}
  (\bibinfo {year} {2011})}\BibitemShut {NoStop}%
\bibitem [{\citenamefont {Backes}\ \emph {et~al.}(2011)\citenamefont {Backes},
  \citenamefont {Casanova},\ and\ \citenamefont {Bruno}}]{Backes2011}%
  \BibitemOpen
  \bibfield  {author} {\bibinfo {author} {\bibfnamefont {A.~R.}\ \bibnamefont
  {Backes}}, \bibinfo {author} {\bibfnamefont {D.}~\bibnamefont {Casanova}}, \
  and\ \bibinfo {author} {\bibfnamefont {O.~M.}\ \bibnamefont {Bruno}},\
  }\bibfield  {title} {\enquote {\bibinfo {title} {IdentificaÁ„o de plantas
  por an·lise de textura foliar},}\ }\href@noop {} {\bibfield  {journal}
  {\bibinfo  {journal} {Learning and Nonlinear Models},\ }\textbf {\bibinfo
  {volume} {9}},\ \bibinfo {pages} {84--90} (\bibinfo {year}
  {2011})}\BibitemShut {NoStop}%
\bibitem [{\citenamefont {Galas}\ \emph {et~al.}(1985)\citenamefont {Galas},
  \citenamefont {Eggert},\ and\ \citenamefont {Waterman}}]{GalasEW1985}%
  \BibitemOpen
  \bibfield  {author} {\bibinfo {author} {\bibfnamefont {D.~J.}\ \bibnamefont
  {Galas}}, \bibinfo {author} {\bibfnamefont {M.}~\bibnamefont {Eggert}}, \
  and\ \bibinfo {author} {\bibfnamefont {M.~S.}\ \bibnamefont {Waterman}},\
  }\bibfield  {title} {\enquote {\bibinfo {title} {Rigorous pattern-recognition
  methods for dna sequences: Analysis of promoter sequences from escherichia
  coli},}\ }\Doi {10.1016/0022-2836(85)90262-1} {\bibfield  {journal} {\bibinfo
   {journal} {Journal of Molecular Biology},\ }\textbf {\bibinfo {volume}
  {186}},\ \bibinfo {pages} {117 -- 128} (\bibinfo {year} {1985})},\ ISSN
  \bibinfo {issn} {0022-2836}\BibitemShut {NoStop}%
\bibitem [{\citenamefont {Isasi}\ \emph {et~al.}(2011)\citenamefont {Isasi},
  \citenamefont {Zapirain},\ and\ \citenamefont {Zorrilla}}]{IsasiZZ2011}%
  \BibitemOpen
  \bibfield  {author} {\bibinfo {author} {\bibfnamefont {A.~G.}\ \bibnamefont
  {Isasi}}, \bibinfo {author} {\bibfnamefont {B.~G.}\ \bibnamefont {Zapirain}},
  \ and\ \bibinfo {author} {\bibfnamefont {A.~M.}\ \bibnamefont {Zorrilla}},\
  }\bibfield  {title} {\enquote {\bibinfo {title} {Melanomas non-invasive
  diagnosis application based on the abcd rule and pattern recognition image
  processing algorithms},}\ }\href@noop {} {\bibfield  {journal} {\bibinfo
  {journal} {Computers in Biology and Medicine},\ }\textbf {\bibinfo {volume}
  {41}},\ \bibinfo {pages} {742 -- 755} (\bibinfo {year} {2011})},\ ISSN
  \bibinfo {issn} {0010-4825}\BibitemShut {NoStop}%
\bibitem [{\citenamefont {Landeweerd}\ \emph {et~al.}(1981)\citenamefont
  {Landeweerd}, \citenamefont {Gelsema}, \citenamefont {Bins},\ and\
  \citenamefont {Halie}}]{LandeweerdGBH1981}%
  \BibitemOpen
  \bibfield  {author} {\bibinfo {author} {\bibfnamefont {G.}~\bibnamefont
  {Landeweerd}}, \bibinfo {author} {\bibfnamefont {E.}~\bibnamefont {Gelsema}},
  \bibinfo {author} {\bibfnamefont {M.}~\bibnamefont {Bins}}, \ and\ \bibinfo
  {author} {\bibfnamefont {M.}~\bibnamefont {Halie}},\ }\bibfield  {title}
  {\enquote {\bibinfo {title} {Interactive pattern recognition of blood cells
  in malignant lymphomas},}\ }\href@noop {} {\bibfield  {journal} {\bibinfo
  {journal} {Pattern Recognition},\ }\textbf {\bibinfo {volume} {14}},\
  \bibinfo {pages} {239 -- 244} (\bibinfo {year} {1981})},\ ISSN \bibinfo
  {issn} {0031-3203}\BibitemShut {NoStop}%
\bibitem [{\citenamefont {Este}\ \emph {et~al.}(2009)\citenamefont {Este},
  \citenamefont {Gringoli},\ and\ \citenamefont {Salgarelli}}]{Este2009}%
  \BibitemOpen
  \bibfield  {author} {\bibinfo {author} {\bibfnamefont {A.}~\bibnamefont
  {Este}}, \bibinfo {author} {\bibfnamefont {F.}~\bibnamefont {Gringoli}}, \
  and\ \bibinfo {author} {\bibfnamefont {L.}~\bibnamefont {Salgarelli}},\
  }\bibfield  {title} {\enquote {\bibinfo {title} {Support vector machines for
  tcp traffic classification},}\ }\Doi {10.1016/j.comnet.2009.05.003}
  {\bibfield  {journal} {\bibinfo  {journal} {Comput. Netw.},\ }\textbf
  {\bibinfo {volume} {53}},\ \bibinfo {pages} {2476--2490} (\bibinfo {year}
  {2009})},\ ISSN \bibinfo {issn} {1389-1286}\BibitemShut {NoStop}%
\bibitem [{\citenamefont {Canals}\ \emph {et~al.}(2010)\citenamefont {Canals},
  \citenamefont {Morro},\ and\ \citenamefont {Rossella}}]{CanalsMR2010}%
  \BibitemOpen
  \bibfield  {author} {\bibinfo {author} {\bibfnamefont {V.}~\bibnamefont
  {Canals}}, \bibinfo {author} {\bibfnamefont {A.}~\bibnamefont {Morro}}, \
  and\ \bibinfo {author} {\bibfnamefont {J.~L.}\ \bibnamefont {Rossella}},\
  }\bibfield  {title} {\enquote {\bibinfo {title} {Stochastic-based
  pattern-recognition analysis},}\ }\Doi {10.1016/j.patrec.2010.07.008}
  {\bibfield  {journal} {\bibinfo  {journal} {Pattern Recognition Letters},\
  }\textbf {\bibinfo {volume} {31}},\ \bibinfo {pages} {2353 -- 2356} (\bibinfo
  {year} {2010})},\ ISSN \bibinfo {issn} {0167-8655}\BibitemShut {NoStop}%
\bibitem [{\citenamefont {Koenderink}(1984)}]{Koenderink1984}%
  \BibitemOpen
  \bibfield  {author} {\bibinfo {author} {\bibfnamefont {J.}~\bibnamefont
  {Koenderink}},\ }\bibfield  {title} {\enquote {\bibinfo {title} {The
  structure of images},}\ }\href@noop {} {\bibfield  {journal} {\bibinfo
  {journal} {Biological Cybernetics},\ }\textbf {\bibinfo {volume} {50}},\
  \bibinfo {pages} {363--370} (\bibinfo {year} {1984})},\ ISSN \bibinfo {issn}
  {0340-1200}\BibitemShut {NoStop}%
\bibitem [{\citenamefont {Witkin}(1983)}]{Witkin1983}%
  \BibitemOpen
  \bibfield  {author} {\bibinfo {author} {\bibfnamefont {A.~P.}\ \bibnamefont
  {Witkin}},\ }\bibfield  {title} {\enquote {\bibinfo {title} {Scale-space
  filtering},}\ }in\ \href@noop {} {\emph {\bibinfo {booktitle} {IJCAI}}}\
  (\bibinfo {year} {1983})\ pp.\ \bibinfo {pages} {1019--1022}\BibitemShut
  {NoStop}%
\bibitem [{\citenamefont {Julesz}(1981)}]{Julesz1981}%
  \BibitemOpen
  \bibfield  {author} {\bibinfo {author} {\bibfnamefont {B.}~\bibnamefont
  {Julesz}},\ }\bibfield  {title} {\enquote {\bibinfo {title} {Textons, the
  elements of texture perception, and their interactions},}\ }\Doi
  {10.1038/290091a0} {\bibfield  {journal} {\bibinfo  {journal} {Nature},\
  }\textbf {\bibinfo {volume} {290}},\ \bibinfo {pages} {91--97} (\bibinfo
  {year} {1981})}\BibitemShut {NoStop}%
\bibitem [{\citenamefont {Leung}\ and\ \citenamefont
  {Malik}(1999)}]{Leung1999}%
  \BibitemOpen
  \bibfield  {author} {\bibinfo {author} {\bibfnamefont {T.}~\bibnamefont
  {Leung}}\ and\ \bibinfo {author} {\bibfnamefont {J.}~\bibnamefont {Malik}},\
  }\bibfield  {title} {\enquote {\bibinfo {title} {Recognizing surfaces using
  three-dimensional textons},}\ }in\ \Doi {10.1109/ICCV.1999.790379} {\emph
  {\bibinfo {booktitle} {Computer Vision, 1999. The Proceedings of the Seventh
  IEEE International Conference on}}},\ Vol.~\bibinfo {volume} {2}\ (\bibinfo
  {year} {1999})\ pp.\ \bibinfo {pages} {1010--1017}\BibitemShut {NoStop}%
\bibitem [{\citenamefont {Staley}\ \emph {et~al.}(2004)\citenamefont {Staley},
  \citenamefont {Matta},\ and\ \citenamefont {Waterman}}]{Staley2004}%
  \BibitemOpen
  \bibfield  {author} {\bibinfo {author} {\bibfnamefont {D.~D.}\ \bibnamefont
  {Staley}}, \bibinfo {author} {\bibfnamefont {M.~S.}\ \bibnamefont {Matta}}, \
  and\ \bibinfo {author} {\bibfnamefont {E.~L.}\ \bibnamefont {Waterman}},\
  }\href@noop {} {\emph {\bibinfo {title} {Chemistry}}},\ edited by\ \bibinfo
  {editor} {\bibfnamefont {A.~C.}\ \bibnamefont {Wilbraham}}\ (\bibinfo
  {publisher} {Pearson Prentice Hall},\ \bibinfo {year} {2004})\BibitemShut
  {NoStop}%
\bibitem [{\citenamefont {Haralick}\ \emph {et~al.}(1973)\citenamefont
  {Haralick}, \citenamefont {Shanmugam},\ and\ \citenamefont
  {Dinstein}}]{Haralick1973}%
  \BibitemOpen
  \bibfield  {author} {\bibinfo {author} {\bibfnamefont {R.~M.}\ \bibnamefont
  {Haralick}}, \bibinfo {author} {\bibfnamefont {K.}~\bibnamefont {Shanmugam}},
  \ and\ \bibinfo {author} {\bibfnamefont {I.}~\bibnamefont {Dinstein}},\
  }\bibfield  {title} {\enquote {\bibinfo {title} {Textural features for image
  classification},}\ }\Doi {10.1109/TSMC.1973.4309314} {\bibfield  {journal}
  {\bibinfo  {journal} {IEEE Transactions on Systems, Man and Cybernetics},\
  }\textbf {\bibinfo {volume} {3}},\ \bibinfo {pages} {610--621} (\bibinfo
  {year} {1973})},\ ISSN \bibinfo {issn} {0018-9472}\BibitemShut {NoStop}%
\bibitem [{\citenamefont {Brodatz}(1966)}]{Brodatz1966}%
  \BibitemOpen
  \bibfield  {author} {\bibinfo {author} {\bibfnamefont {P.}~\bibnamefont
  {Brodatz}},\ }\href@noop {} {\emph {\bibinfo {title} {Textures, a
  photographic album for artists and designers}}}\ (\bibinfo  {publisher}
  {Dover Publications New York},\ \bibinfo {year} {1966})\ ISBN \bibinfo {isbn}
  {0486216691},\ p.\ \bibinfo {pages} {112}\BibitemShut {NoStop}%
\bibitem [{vis(2009)}]{vistex2009}%
  \BibitemOpen
  \href {vismod.media.mit.edu/vismod/imagery/VisionTexture/vistex.html}
  {\enquote {\bibinfo {title} {Vision texture database},}\ } (\bibinfo {year}
  {2009})\BibitemShut {NoStop}%
\bibitem [{\citenamefont {Backes}\ \emph {et~al.}(2012)\citenamefont {Backes},
  \citenamefont {Casanova},\ and\ \citenamefont {Bruno}}]{usptex2012}%
  \BibitemOpen
  \bibfield  {author} {\bibinfo {author} {\bibfnamefont {A.~R.}\ \bibnamefont
  {Backes}}, \bibinfo {author} {\bibfnamefont {D.}~\bibnamefont {Casanova}}, \
  and\ \bibinfo {author} {\bibfnamefont {O.~M.}\ \bibnamefont {Bruno}},\
  }\bibfield  {title} {\enquote {\bibinfo {title} {Color texture analysis based
  on fractal descriptors},}\ }\Doi {10.1016/j.patcog.2011.11.009} {\bibfield
  {journal} {\bibinfo  {journal} {Pattern Recognition},\ }\textbf {\bibinfo
  {volume} {45}},\ \bibinfo {pages} {1984 -- 1992} (\bibinfo {year} {2012})},\
  ISSN \bibinfo {issn} {0031-3203}\BibitemShut {NoStop}%
\bibitem [{\citenamefont {Ojala}\ \emph {et~al.}(2002)\citenamefont {Ojala},
  \citenamefont {M\"aenp\"a\"a}, \citenamefont {Pietik\"ainen}, \citenamefont
  {Viertola}, \citenamefont {Kyll\"{o}nen},\ and\ \citenamefont
  {Huovinen}}]{Ojala2002}%
  \BibitemOpen
  \bibfield  {author} {\bibinfo {author} {\bibfnamefont {T.}~\bibnamefont
  {Ojala}}, \bibinfo {author} {\bibfnamefont {T.}~\bibnamefont
  {M\"aenp\"a\"a}}, \bibinfo {author} {\bibfnamefont {M.}~\bibnamefont
  {Pietik\"ainen}}, \bibinfo {author} {\bibfnamefont {J.}~\bibnamefont
  {Viertola}}, \bibinfo {author} {\bibfnamefont {J.}~\bibnamefont
  {Kyll\"{o}nen}}, \ and\ \bibinfo {author} {\bibfnamefont {S.}~\bibnamefont
  {Huovinen}},\ }\bibfield  {title} {\enquote {\bibinfo {title} {Outex - new
  framework for empirical evaluation of texture analysis algorithms},}\ }in\
  \href@noop {} {\emph {\bibinfo {booktitle} {Proceedings of the 16 th
  International Conference on Pattern Recognition (ICPR 2002)}}},\ \bibinfo
  {series} {ICPR 2002}, Vol.~\bibinfo {volume} {1}\ (\bibinfo  {publisher}
  {IEEE Computer Society},\ \bibinfo {year} {2002})\ p.\ \bibinfo {pages}
  {10701}\BibitemShut {NoStop}%
\bibitem [{\citenamefont {P\'eteri}\ \emph {et~al.}(2010)\citenamefont
  {P\'eteri}, \citenamefont {Fazekas},\ and\ \citenamefont
  {Huiskes}}]{DyntexPeteri2010}%
  \BibitemOpen
  \bibfield  {author} {\bibinfo {author} {\bibfnamefont {R.}~\bibnamefont
  {P\'eteri}}, \bibinfo {author} {\bibfnamefont {S.}~\bibnamefont {Fazekas}}, \
  and\ \bibinfo {author} {\bibfnamefont {M.~J.}\ \bibnamefont {Huiskes}},\
  }\bibfield  {title} {\enquote {\bibinfo {title} {{D}yn{T}ex : a
  {C}omprehensive {D}atabase of {D}ynamic {T}extures},}\ }\href@noop {}
  {\bibfield  {journal} {\bibinfo  {journal} {Pattern Recognition Letters},\
  }\textbf {\bibinfo {volume} {31}},\ \bibinfo {pages} {1627--1632} (\bibinfo
  {year} {2010})},\ \bibinfo {note}
  {http://projects.cwi.nl/dyntex/}\BibitemShut {NoStop}%
\end{thebibliography}

%

\end{document}